\documentclass[sigconf]{acmart}

\usepackage{algorithmic}
\usepackage{siunitx}
\usepackage{booktabs}       % professional-quality tables
\usepackage{xcolor}         % colors
\usepackage{graphicx}
\usepackage{bm}
\usepackage{caption}
\usepackage{subcaption}
\usepackage{pifont}% http://ctan.org/pkg/pifont
\usepackage{multirow} 
\usepackage{xspace}
\usepackage{float}

\definecolor{BrickRed}{RGB}{170, 74, 68}
\definecolor{ForestGreen}{RGB}{34, 139, 34}
\AtBeginDocument{%
  \providecommand\BibTeX{{%
    \normalfont B\kern-0.5em{\scshape i\kern-0.25em b}\kern-0.8em\TeX}}}

\acmConference[Conference acronym 'XX]{Make sure to enter the correct
  conference title from your rights confirmation emai}{June 03--05,
  2018}{Woodstock, NY}
\acmPrice{15.00}
\acmISBN{978-1-4503-XXXX-X/18/06}

\newcommand{\specialcell}[2][l]{\begin{tabular}[#1]{@{}l@{}}#2\end{tabular}}
\newcommand{\model}{COSMO\xspace} 
\newcommand{\dataset}{iWildCam2020-WILDS\xspace} 
\newcommand{\cmark}{\ding{51}}
\newcommand\eg[0]{\textit{e.g.}}
\newcommand\ie[0]{\textit{i.e.}}

\title{Reviving the Context: Camera Trap Species Classification as Link Prediction on Multimodal Knowledge Graphs}

\author{Vardaan Pahuja}
\email{pahuja.9@osu.edu}
\affiliation{%
  \institution{The Ohio State University}
  \city{Columbus}
  \state{Ohio}
  \country{United States}
}

\author{Weidi Luo}
\email{luo.1455@osu.edu}
\affiliation{%
  \institution{The Ohio State University}
  \city{Columbus}
  \state{Ohio}
  \country{United States}
}

\author{Yu Gu}
\email{gu.826@osu.edu}
\affiliation{%
  \institution{The Ohio State University}
  \city{Columbus}
  \state{Ohio}
  \country{United States}
}

\author{Cheng-Hao Tu}
\email{tu.343@osu.edu}
\affiliation{%
  \institution{The Ohio State University}
  \city{Columbus}
  \state{Ohio}
  \country{United States}
}

\author{Hong-You Chen}
\email{chen.9301@osu.edu}
\affiliation{%
  \institution{The Ohio State University}
  \city{Columbus}
  \state{Ohio}
  \country{United States}
}

\author{Tanya Berger-Wolf}
\email{berger-wolf.1@osu.edu}
\affiliation{%
  \institution{The Ohio State University}
  \city{Columbus}
  \state{Ohio}
  \country{United States}
}

\author{Charles Stewart}
\email{stewart@rpi.edu}
\affiliation{%
  \institution{Rensselaer Polytechnic Institute}
  \city{Troy}
  \state{New York}
  \country{United States}
}

\author{Song Gao}
\email{song.gao@wisc.edu}
\affiliation{%
  \institution{University of Wisconsin-Madison}
  \city{Madison}
  \state{Wisconsin}
  \country{United States}
}

\author{Wei-Lun Chao}
\email{chao.209@osu.edu}
\affiliation{%
  \institution{The Ohio State University}
  \city{Columbus}
  \state{Ohio}
  \country{United States}
}

\author{Yu Su}
\email{su.809@osu.edu}
\affiliation{%
  \institution{The Ohio State University}
  \city{Columbus}
  \state{Ohio}
  \country{United States}
}

\begin{document}

\begin{abstract}
  Camera traps are important tools in animal ecology for biodiversity monitoring and conservation.
  However, their practical application is limited by issues such as poor generalization to new and unseen locations.
  Images are typically associated with diverse forms of context, which may exist in different modalities.
  In this work, we exploit the structured context linked to camera trap images to boost out-of-distribution generalization for species classification tasks in camera traps.
  For instance, a picture of a wild animal could be linked to details about the time and place it was captured, as well as structured biological knowledge about the animal species. 
  While often overlooked by existing studies, incorporating such context offers several potential benefits for better image understanding, such as addressing data scarcity and enhancing generalization.
  However, effectively incorporating such heterogeneous context into the visual domain is a challenging problem.
  To address this, we propose a novel framework that transforms species classification as link prediction in a multimodal knowledge graph (KG).
  This framework enables the seamless integration of diverse multimodal contexts for visual recognition.
  We apply this framework for out-of-distribution species classification on the iWildCam2020-WILDS and Snapshot Mountain Zebra datasets and achieve competitive performance with state-of-the-art approaches. Furthermore, our framework enhances sample efficiency for recognizing under-represented species.\footnote{Our code is available at \url{https://github.com/OSU-NLP-Group/COSMO}}
  % {\color{blue} revise it}
\end{abstract}




\maketitle
\renewcommand{\shortauthors}{Vardaan Pahuja et al.}

\section{Introduction}
Human activities are increasingly endangering wildlife species, resulting in a significant global decline in animal populations \cite{maxwell2016biodiversity, diaz2019pervasive, almond2020living}. 
% As a result, there is a crucial need to identify, monitor, and quantify the distribution of wildlife species in order to actively preserve ecological biodiversity. 
Therefore, accurately identifying and tracking wildlife species is vital for preserving ecological biodiversity.
% To streamline data collection and reduce logistical expenses, ecologists deploy camera traps --- digital cameras triggered by motion or infrared radiation in natural habitats \cite{o2011camera, wearn2017camera, glover2019camera}. 
Camera traps, digital cameras activated by motion or infrared in natural habitats, have become ecologists' preferred data collection tool \cite{o2011camera, wearn2017camera, glover2019camera}.
% Recognizing animal species by domain experts is a time-consuming and laborious process, which is not practical for the scale of data generated by these camera traps. 
% Thus, computer vision approaches have been leveraged in the literature to automatically recognize animal species in camera trap images \cite{weinstein2018computer, sadeghnorouzzadeh2020a, ahumada2020wildlife, DBLP:conf/icml/KohSMXZBHYPGLDS21, DBLP:journals/corr/abs-2303-15823}. 
However, manually sifting through the numerous images they capture is a time-consuming and arduous task for experts. 
This has led to the increased use of computer vision techniques for species recognition \cite{DBLP:conf/crv/SchneiderTK18, weinstein2018computer, sadeghnorouzzadeh2020a, ahumada2020wildlife, DBLP:conf/icml/KohSMXZBHYPGLDS21, DBLP:journals/corr/abs-2303-15823}. 
Yet, a challenge has arisen: many of these models overfit to the backgrounds of their training images, diminishing their effectiveness on images from new locations \cite{beery2018recognition, miao2019insights, schneider2020three}. 
This underscores the need for \emph{more adaptable species classification models that perform well across diverse \textbf{contexts}}.

% Extensive research in cognitive science has elucidated the pivotal significance of contextual information in shaping human perception and facilitating the process of visual recognition~\cite{bar2004visual,oliva2007role,aminoff2007parahippocampal}. 
Building on this, cognitive science research has demonstrated the profound influence of \textbf{\emph{contextual}} information on human perception and visual recognition processes~\cite{bar2004visual,oliva2007role,aminoff2007parahippocampal}. 
% In many real-world scenarios, images---a common visual signal---are invariably embedded within a multitude of diverse contextual elements.
% For instance, images shared by users on social media platforms 
% like Flickr~\cite{10.1145/1460096.1460104} 
% are accompanied by links to the profiles of the individuals who posted or liked them, along with relevant tags.
% Similarly, on e-commerce websites, product images are often surrounded by a plethora of metadata, including price, warranty, color, and more.
% Photos of wildlife taken by camera traps also involve heterogeneous metadata, such as where (\ie, camera location coordinates) and when (\ie, timestamps) a photo is taken.
Particularly in wildlife monitoring, camera trap images are replete with crucial contextual data, such as where (\ie, camera location coordinates) and when (\ie, timestamps) a photo is taken.
Furthermore, the structured knowledge of biology taxonomy (\eg, Open Tree Taxonomy~\cite{OTT}) can also provide valuable context for understanding the species in camera trap images.
Such context provides important knowledge that can boost the recognition of visual concepts. 
For instance, the knowledge that a certain feline image was taken from a camera trap in Africa significantly reduces the likelihood of it representing a tiger.
% For example, we may infer that a feline in the photo is unlikely to be a tiger if the photo is taken in Africa. 
% Similarly, if an image is being shown on Amazon, we can discern that it is more likely a stuffed toy than a live rabbit.
In addition, more robust associations might be learned with the aid of contextual information because the context provides invariable knowledge that is unbiased towards variations in the illuminations or angles of an image.
This may help to compensate for domain shifts in species images resulting from such variations and potentially lead to better out-of-distribution (OOD) generalizability~\cite{bargoti2016image,ellen2019improving}.
Consequently, the incorporation of contextual information in species identification presents a significant problem worthy of investigation.

Nevertheless, contextual information has been under-exploited in the literature of image classification; standard image classification models~\cite{DBLP:journals/corr/SimonyanZ14a, he2016deep} often disregard the contextual information tied to images.
% This is partly due to the difficulty of incorporating contextual information in image classification with a unified learning framework due to the heterogeneous nature of the context.
This is partly due to the heterogeneous nature of the context, which makes it challenging to incorporate contextual information in image classification using a unified learning framework.
Contextual information in different modalities (\eg, numerical values, textual descriptions, or structured taxonomies) is usually represented separately from the image in \textit{distinct feature spaces}. 
The question of effectively combining features from these different spaces within a unified learning framework remains unanswered.
Existing research typically treats all the features as additional input to the classifier via feature vector concatenation~\cite{li2014exploiting, bargoti2016image, ellen2019improving} or utilizes fusion to obtain aggregate representations \cite{chu2019geo, diao2022metaformer}.
% Such methods, despite their simplicity, cannot capture complex structural and semantic relationships between images and various contextual information.
Despite their simplicity, such approaches are incapable of capturing complex structural and semantic relationships between images and various contextual information.
Additionally, these approaches assume a uniform availability of contextual information across all images, which is often unrealistic in real-world scenarios.
As a result, their flexibility is limited, especially when considering situations where certain images may lack some contextual details, such as coordinates or timestamps, like in camera trap photos.

\begin{figure*}[htbp]
    \centering
    \includegraphics[width=0.8\linewidth]{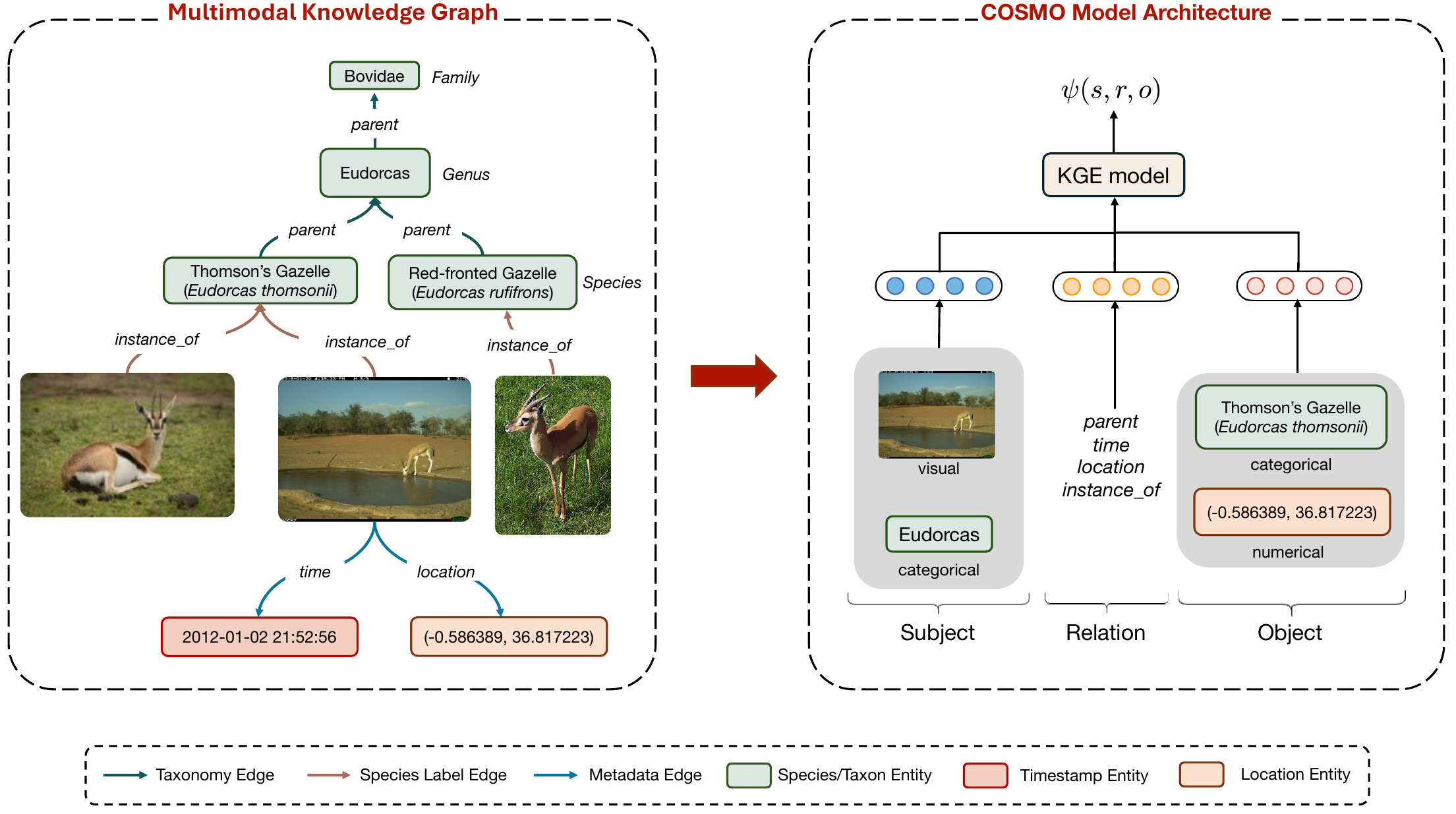}
    \caption{\textbf{Overview of our framework \model.} \textbf{\textit{Left:}} Our multimodal knowledge graph for camera traps and wildlife. Photos from camera traps are jointly represented in the KG with contextual information such as time, location, and structured biology taxonomy. The taxonomy is obtained from Open Tree Taxonomy (OTT)~\cite{OTT} or iNaturalist \cite{van2018inaturalist}. \textbf{\textit{Right:}} In our formulation of species classification as link prediction, the plausibility score $\psi(s, r, o)$ of each (subject, relation, object) triplet is computed using a KGE model (\eg, DistMult), where the subject, relation, and object are all first embedded into a vector space. Specifically, for our multimodal KG, we represent visual entities using a ResNet-50 pre-trained on ImageNet and represent numerical entities using an MLP. For categorical entities and relations, we directly represent them with embedding lookups.}
    \label{fig:mmkg_model}
\end{figure*}

Towards this end, we propose a new learning framework, \model (\underline{C}lassification \underline{O}f \underline{S}pecies using \underline{M}ultimodal c\underline{O}ntext), where we first organize all species images and contextual information as a \textit{multimodal knowledge graph} (KG) and then reformulate species classification as the standard link prediction task on the KG.
Specifically, we consider species images, their corresponding labels (which are available in the training data), and their associated attributes provided in the context as entities within our KG (see Figure~\ref{fig:mmkg_model} for an example).
We represent the relationships between these entities as edges in our KG (see a more concrete description of our KG construction in Section~\ref{sec:kbc}). 
Our KG is multimodal because its entities belong to different modalities.
% Our KG is multimodal since its entities are from different modalities.
In this context, species classification can be framed as a link prediction task, where the objective is to predict the presence of an edge between an image and its corresponding species label within the KG.
This learning framework enables a unified way to incorporate heterogeneous contextual information for species classification. 
Each form of multimodal information is treated as a type of entity, a first-class citizen of the multimodal KG with its representation computed using a modality-specific encoder. 
The learning process enables the interaction of different modalities in a \textit{joint feature space} for robust representation learning.
% and uses relational knowledge to regularize their features. 
% This regularization mechanism holds the potential to extract robust patterns for image classification through KG reasoning, thereby enhancing OOD generalization. 
In addition, \model demonstrates greater flexibility by not assuming uniform availability of all contextual information, unlike previous methods.

% \begin{figure}
%     \centering
%     \includegraphics[width=0.65\linewidth]{figures/intro_fig_new.pdf}
%     \caption{A wildlife animal image associated with the heterogeneous multimodal context.}
%     \label{fig:intro_img}
% \end{figure}

We employ the widely used DistMult \cite{DistMult} model as our backbone model for link prediction to instantiate the \model framework. 
To assess the performance of \model, particularly in terms of out-of-distribution generalization, we conduct experiments on the iWildCam2020-WILDS benchmark~\cite{DBLP:conf/icml/KohSMXZBHYPGLDS21} and Snapshot Mountain Zebra \cite{pardo2021snapshot}, which are standard datasets for species classification in camera trap photos.
They contain naturally occurring wildlife photos associated with metadata. Factors like variation in illumination, camera pose, and motion blur pose challenges for robustness and generalization, making these benchmarks an ideal testbed for assessing our framework's effectiveness. 
% During training, our framework learns multimodal KG representations. During inference, it solely relies on the image to make predictions. 
% In the no-context setting which only uses the image-species labels, our model performs on par with the standard image classification baselines, which is quite impressive given the additional complexity to learn relation representations. 
We show that \model offers a unified framework to incorporate heterogeneous context leading to improved species classification performance over existing out-of-distribution generalization approaches.

The main contribution of this work is three-fold:
\begin{itemize}
    \item We propose a novel framework, \model, that reformulates species classification as link prediction in a multimodal knowledge graph, which provides a unified way to incorporate heterogeneous forms of contextual information associated with images for visual recognition.
    
    \item We instantiate this framework for species classification of wildlife images, including the construction of a novel multimodal KG for this problem that integrates spatiotemporal information and structured biology knowledge. 
    
    \item Evaluation on the standard iWildCam2020-WILDS and Snapshot Mountain Zebra datasets demonstrate that \model achieves competitive performance compared with standard species classification methods, especially in improving robustness and OOD generalization.
    % leads to robust OOD generalization to unseen domains and improves sample efficiency for recognizing under-represented species examples.
\end{itemize}

% \FloatBarrier

\section{Related Work}

\noindent \textbf{Species Recognition in Camera Traps.}
Deep neural networks such as CNNs have been successfully deployed for large-scale recognition of camera trap images \cite{weinstein2018computer, norouzzadeh2018automatically, tabak2019machine}. This has paved the way for significant savings in logistics costs for biodiversity conservation. However, training such models often requires enormous amounts of data to perform well. \citeauthor{sadeghnorouzzadeh2020a} and \citeauthor{DBLP:journals/corr/abs-2303-15823} propose active learning approaches to mitigate the sample inefficiency of training species classification models in such systems.
Another challenge arises from the tendency of these models to overfit to the backgrounds present in the training images \cite{beery2018recognition, miao2019insights}, which limits their deployment to new camera trap locations \cite{tabak2019machine, schneider2020three}. Improving robustness to new locations is a significant research challenge \cite{beery2019iwildcam, tabak2020improving} leading to the curation of datasets like iWildCam2020-WILDS \cite{DBLP:conf/icml/KohSMXZBHYPGLDS21} to test OOD generalization for such systems.
Domain adaptation approaches in the literature seek to mitigate this issue by distributionally robust optimization \cite{DBLP:conf/icml/HuNSS18, qi2020attentional} or learning domain invariant features \cite{DBLP:conf/eccv/SunS16}.
In contrast, this work helps improve the robustness to new camera trap locations by utilizing a multimodal KG of heterogeneous contexts.\\

\noindent \textbf{Image Classification with Auxiliary Information.}
Despite the ubiquity of contextual metadata, the potential of leveraging them for image classification has been largely under-explored.
Previous studies have primarily treated metadata as additional input features for classifiers~\cite{li2014exploiting, bargoti2016image, ellen2019improving}, representing a shallow use that fails to capture the intricate relationships between metadata and images. 
Some works have attempted to model pairwise dependencies between images using heuristics based on metadata, such as shared tags on social media~\cite{mcauley2012image, long2019deep} or aggregating information from neighborhood images with similar metadata \cite{johnson2015love} while disregarding more complex relationships among images, metadata, and labels.
% Geo-Aware Networks \cite{chu2019geo} proposes an additive fusion of location-based metadata features and visual features for visual recognition.
Metaformer \cite{diao2022metaformer} feeds a sequence of image patches and metadata to a Transformer model for their fusion.
Additionally, these methods assume a uniform availability of metadata for all images, which is often not the case in reality due to data scarcity.
For instance, the camera trap location coordinates may not be available in some cases due to privacy and security reasons.
In our work, we do not assume such uniform availability and build the multimodal KG using available metadata.

Apart from the metadata, external sources of knowledge are also used in image classification.
For instance, ~\citeauthor{jayathilaka2021ontology} embed each class as a vector based on a hierarchy derived from WordNet~\cite{miller1995wordnet}.
\citeauthor{DBLP:journals/tvcg/AlsallakhJY0R18} develop a class hierarchy-aware CNN for image classification on ImageNet. Similarly, \citeauthor{bertinetto2020making} and \citeauthor{DBLP:conf/cvpr/ZhangXXR22} design hierarchy-aware objectives to incorporate taxonomy in image representations.
BioCLIP \cite{stevens2023bioclip} verbalizes the taxonomic hierarchy to train a CLIP-style foundational model for species classification across plants, animals, and fungi.
\citeauthor{marino2017more} represent images as local subgraphs of Visual Genome~\cite{krishna2017visual}.
% with the aid of an object detector.
In contrast, \model constructs a global KG with both metadata and external knowledge, \eg, taxonomy information from Open Tree Taxonomy, and approaches image classification as link prediction within the KG.\@
Our novel formulation is flexible in handling data scarcity of metadata and enables reasoning over diverse relationships present in the KG.\\

\noindent \textbf{KG Link Prediction.}
Most real-world KGs are incomplete. The task of link prediction or knowledge graph completion (KGC) tries to infer missing links given the observed ones. Early approaches for link prediction range from translation-based models \cite{bordes2013translating, lin2015learning} and semantic matching models \cite{nickel2011three, DistMult} to the ones that leverage neural networks like feedforward neural networks \cite{dong2014knowledge}, CNNs \cite{dettmers2018convolutional, nguyen2017novel}, and Transformer-based models \cite{Yao2019KGBERTBF, chen-etal-2021-hitter, saxena-etal-2022-sequence}.
These methods use a parameterized scoring function based on learned entity and relation embeddings to calculate the plausibility of a particular triplet. 
However, it could be challenging to fully encode the rich semantic information of KGs into such shallow embeddings.
To mitigate this, \citeauthor{schlichtkrull2018modeling}, \citeauthor{vashishth2020compositionbased}, \citeauthor{DBLP:conf/www/YuYZW21}, \citeauthor{10.1145/3583780.3614769} use graph neural networks (GNNs) to encode the rich neighborhood context of entities for link prediction. In our framework, we employ a global multimodal KG, which consists of biological taxonomy and metadata, as the context to enhance OOD generalization.\\

\noindent \textbf{Multimodal KG Reasoning.}
Multimodal KGs extend traditional KGs by including entities of different modalities such as categorical data, images, numerical data, etc.
KBLRN \cite{garcia2017kblrn} is a pioneering work in multimodal KG reasoning that uses extra information in the form of relational and numerical features for multimodal KG reasoning. Similarly, IKRL \cite{DBLP:conf/starsem/SergiehBGR18} proposes a fusion of linguistic and visual information with structured information for link prediction.
MKBE \cite{pezeshkpour-etal-2018-embedding} constructs a multimodal KG using numerical, image, and textual information, treating them as entities instead of auxiliary features, for the link prediction task.
MR-GCN \cite{wilcke2023} further extends it by including support for more modalities, \eg, numerical, temporal, textual, visual, and spatial predicate links in the multimodal KG\@. To provide a more expressive way for interaction between different modalities, IMF \cite{li2023imf} uses bilinear pooling to fuse multiple modality features and trains it using contrastive learning on the contextual entity representations. 
Our work leverages link prediction in a multimodal KG to enable out-of-distribution generalization for species classification in camera traps.

% \FloatBarrier

\section{Methodology}

% {\color{blue} add preliminaries }
\subsection{Preliminaries}
\noindent \textbf{Multimodal KG.} Given a set of KG entities with categorical values $\mathcal{E}_{\mathcal{KG}}$, multimodal entities $\mathcal{E}_{\mathcal{MM}}$, and a set of relations $\mathcal{R}$, a multimodal KG can be defined as a collection of facts $\mathcal{F} \subseteq (\mathcal{E}_{\mathcal{KG}}\cup\mathcal{E}_{\mathcal{MM}}) \times \mathcal{R} \times (\mathcal{E}_{\mathcal{KG}}\cup\mathcal{E}_{\mathcal{MM}})$ where for each fact $f = (h, r, t)$, $\;h,t \in (\mathcal{E}_{\mathcal{KG}}\cup\mathcal{E}_{\mathcal{MM}}), r \in \mathcal{R}$.

\noindent \textbf{KG Link Prediction.} The task of link prediction is to infer missing facts based on known facts in a KG\@. Given a link prediction query $(h, r, ?)$ or $(?, r, t)$, the model ranks the target entity among the set of candidate entities.
% The model is trained using a contrastive loss to maximize the score of the observed positive triples compared to randomly sampled negatives.

\noindent \textbf{Problem Setup.}
The task entails species recognition for camera trap images amidst distribution shifts.
% Given an image obtained from a camera trap, the task is to recognize the animal species in that image.
% We use the iWildCam2020-WILDS dataset where the label space has 182 species classes.
The training and test sets comprise images obtained from disjoint camera traps, enabling the evaluation of out-of-domain (OOD) generalization.
During training, we use the multimodal KG to train our model, while we use just the image to make predictions for inference. The goal is to learn visual representations robust to distribution shifts by leveraging the rich structural and semantic information provided by the multimodal knowledge graph.

% {\color{blue} mention train/test input}

% {\color{blue} Describe KG construction}
\subsection{Building the Multimodal KG}\label{sec:kbc}
The multimodal KG comprises entities from different modalities interconnected by heterogeneous relationships. The base KG consists of camera trap images linked with their species labels from the training set (\texttt{<image>}, \texttt{instance of}, \texttt{<species label>}). Next, we progressively augment the KG with links connecting the existing entities to contextual information. In this work, we utilize the following attributes to provide context for species classification:
\begin{itemize}
    \item \textbf{Taxonomy}: The taxonomy forms the core of the multimodal knowledge graph, connecting distinct species to higher-order taxa. For \dataset, we obtain the phylogenetic taxonomy corresponding to the species of interest from Open Tree Taxonomy (OTT) ~\cite{OTT} and manually link it to the species in the dataset.
    % It contains evolutionary relationships among 2.6M taxa that are organized based on evolutionary relatedness.
    For the Snapshot Mountain Zebra dataset, we utilize the iNaturalist taxonomy \cite{van2018inaturalist} mapping provided by \url{www.lila.science}.
    % {\color{blue} move it to dataset section}
    % Out of 182 wildlife species present in iWildCam2020-WILDS, 155 species are covered in OTT\@. {\color{blue} add more on this}
    
    \item \textbf{Location}: The camera trap images are associated with the GPS coordinates of their source cameras. For the \dataset dataset, this metadata is available for a portion of the images (67\%) and is obfuscated within 1 km.\ for privacy reasons. Animals demonstrate a preference for particular habitats; thus, the location context attribute is useful for species recognition.
    
    \item \textbf{Time}: The timestamp attribute indicates the precise moment when the image was captured. This timestamp information proves valuable in species recognition since specific animals exhibit activity patterns tied to particular times of the day, such as feeding, hunting, or defending their territory. In our multimodal knowledge graph, we utilize the timestamp information at an hourly granularity.
\end{itemize}
% A schematic representation of the multimodal KG for our task is shown in Figure~\ref{fig:mmkg}.
Figure~\ref{fig:mmkg_model} presents a schematic representation of various contexts in a multimodal KG.\@
% We refer to the multimodal KG obtained using training images linked to their species labels (\texttt{<image>}, \texttt{instance of}, \texttt{<species label>}) as the base KG\@.
For location, time, and taxonomy attributes, the corresponding RDF triplets can be represented as (\texttt{<image>}, \texttt{location}, \texttt{<GPS co-ordinate>}), (\texttt{<image>}, \texttt{time}, \texttt{<timestamp>}), and (\texttt{<taxon\textunderscore1>}, \texttt{parent}, \texttt{<taxon\textunderscore2>)}, respectively.

% \begin{figure}
%     \centering
%     \includegraphics[width=0.85\linewidth]{figures/mmkg.pdf}
%     \caption{Multimodal Knowledge Graph for wildlife species}
%     \label{fig:mmkg}
% \end{figure}

% \input{tables/main_table_iwildcam}

\subsection{Model Architecture}
% {\color{blue} Describe model architecture including DistMult}\\
We use DistMult \cite{DistMult}, a strong baseline on KGE benchmarks, as our backbone KG embedding model.\footnote{Recent work \cite{Ruffinelli2020You} showed that simple baselines like DistMult outperform more sophisticated neural network baselines when trained properly.} Note that \model is a general framework that can leverage a variety of KG embedding models proposed in the literature. DistMult minimizes a bilinear scoring function between the entity embeddings of the subject and object entities. For a given triplet $(h, r, t)$, the scoring function of DistMult is defined as:
\begin{equation}
\psi(h, r, t) = \bm{h}^T \bm{W}_r \bm{t} = \sum_{i=1}^{d}\bm{h}_i \cdot diag(\bm{W}_r)_i \cdot \bm{t}_i
\end{equation}
Here, $\bm{h}$ and $\bm{t}$ denote the vector representations of the head entity and tail entity, respectively.
The relation representation is parameterized by $\bm{W}_r \in \mathbb{R}^{d\times d}$, a diagonal matrix.

% {\color{blue} Add image, location, and time encoders and respective loss functions}\\
\subsubsection{Multi-modality Encoders}
We use an ImageNet pre-trained ResNet-50 \cite{he2016deep} as the image encoder. The base feature of each location is represented as a 2D vector \texttt{[latitude, longitude]}. Following prior work \cite{pezeshkpour-etal-2018-embedding}, we use an MLP to project the 2D location feature to a higher dimensional space. Similarly, for temporal context, we use an MLP to project the integer value of the hour timestamp to the higher dimensional embedding space. For categorical entities such as species labels and taxa, we learn dense embeddings as representations.

\subsubsection{Training}
We train the model using an optimization strategy based on the modality of the tail entity. For categorical attributes, we formulate it as a multi-class classification problem and use standard cross-entropy loss to train the model. For instance, in case of a given image-species label ground truth triplet $(\mathcal{I}, \texttt{instance\;of}, s)$, the loss is defined as:
\begin{equation}
\begin{split}
  \mathcal{L}(\mathcal{I}, &\texttt{instance\;of}, s) =\\
  &-\log \frac{\exp(\psi(\mathcal{I}, \texttt{instance\;of}, s))}{\sum_{s^{'}\in S} \exp(\psi(\mathcal{I}, \texttt{instance\;of}, s^{'}))},
\end{split}
\end{equation}
where $S$ denotes the set of all species labels and $\psi(h, r, t)$ denotes the plausibility score of KG edge $(h, r, t)$.

For numerical attributes such as location and time, we formulate it as a multi-class multi-label classification problem and use a binary cross-entropy loss to optimize the parameters. This choice is motivated by the fact that images can be associated with a range of GPS coordinates and timestamps, \eg, most animals are active multiple times during the day.
The label space comprises all entities of ground truth modality. For instance, in the case of a given time modality ground truth triplet $(\mathcal{I}, \texttt{time}, t)$, the loss is defined as:
\begin{equation}
\begin{split}
\mathcal{L}(\mathcal{I}, \texttt{time}, t) = -\sum_{t^{'}} l^{\mathcal{I}, time}_{t^{'}} \cdot \log(\sigma(\psi(\mathcal{I}, \texttt{time}, t^{'}))) +\\
(1 -l^{\mathcal{I}, time}_{t^{'}}) \cdot (1-\log(\sigma(\psi(\mathcal{I}, \texttt{time}, t^{'})))),
\end{split}
\end{equation}
where $l^{\mathcal{I}, time}_{t^{'}}$ is a binary label that indicates whether the triplet $(\mathcal{I}, time, t^{'})$ exists in the set of observed triplets and $\sigma(\cdot)$ is the sigmoid activation function. We train the model by sequentially minimizing the objective on each type of context triplet.
Figure~\ref{fig:mmkg_model} illustrates the overall model architecture.

% {\color{blue} TODO: Add that training is done sequentially}

% \FloatBarrier

\section{Experimental Setup}

\subsection{Datasets}
% {\color{blue} Describe attributes.}\\
We test our approach on the iWildCam2020-WILDS dataset \cite{DBLP:conf/icml/KohSMXZBHYPGLDS21}, a variant of the iWildCam 2020 dataset \cite{iwildcam2020} and Snapshot Mountain Zebra \cite{pardo2021snapshot}.
iWildCam2020-WILDS is a benchmark dataset designed to test OOD generalization for the task of species classification. 
% It consists of wildlife images collected from camera traps, heat or motion-activated cameras placed in the wild \cite{wearn2017camera}.
The label space consists of 182 species.  Each domain corresponds to a different location of the camera trap. The training and test images belong to disjoint sets of locations in the OOD setting. 

\begin{table*}[h]
\centering
\caption{Species classification results on iWildCam2020-WILDS (OOD) dataset.
The first baseline in the second section shows the no-context baseline that uses only image-species labels as KG edges. All models use a pre-trained ResNet-50 as image encoder. Parentheses show standard deviation across 3 random seeds. We highlight the best result in bold and the second best with underline. We mark the improvements over \model (no-context) in {\color{ForestGreen}green}. Missing values are denoted by --.}
\small
\resizebox{\linewidth}{!}{%
\begin{tabular}{cccccll} \toprule
& \multirow{2}{*}{\bfseries Model}               &  \multicolumn{3}{c}{\bfseries Multi-modality}  &                    \multirow{2}{*}{\bfseries Val.\ Acc. (\%)} & \multirow{2}{*}{\bfseries Test Acc.\ (\%) }                   \\ \cmidrule(r){3-5}
&   &  \bfseries Taxonomy & \bfseries Location & \bfseries Time                  &                      &                     \\ \cmidrule(r){1-2} \cmidrule(r){3-5} \cmidrule(r){6-7}
& Empirical Risk Minimization (ERM) \cite{DBLP:conf/icml/KohSMXZBHYPGLDS21}                &  & \multirow{5}{*}{--} &  & 62.7 {\scriptsize($\pm$2.4)}       & 71.6 {\scriptsize($\pm$2.5)}      \\
& CORAL \cite{DBLP:conf/eccv/SunS16} &  &  &             & 60.3 {\scriptsize($\pm$2.8)}      & 73.3 {\scriptsize($\pm$4.3)}      \\
& Group DRO \cite{DBLP:conf/icml/HuNSS18} &  &  &         & 60.0 {\scriptsize($\pm$0.7)}       & 72.7 {\scriptsize($\pm$2.0)}    \\
& Fish \cite{shi2022gradient} &  &  &             & 58.0 {\scriptsize($\pm$0.2)}              &  63.2 {\scriptsize($\pm$0.7)}  \\ 
& ABSGD \cite{qi2020attentional} &  &  &                         &  --             & 72.7 {\scriptsize($\pm$1.8)}   \\
& MLP-concat &  & \cmark & \cmark & 27.3 {\scriptsize($\pm$0.8)} & 39.6 {\scriptsize($\pm$1.0)}\\ \midrule
& \model(no-context) &          & --  &  &  63.2 {\scriptsize($\pm$0.4)}  & 68.8 {\scriptsize($\pm$2.1)} \\ \midrule
\parbox[t]{2mm}{\multirow{3}{*}{\rotatebox[origin=c]{90}{\footnotesize \specialcell{Single\\context\\}}}} & \multirow{3}{*}{\textbf{\model}} & \cmark &  &  & 62.8 {\scriptsize($\pm$2.2)} ({\color{BrickRed}-0.4})  & 72.4 {\scriptsize($\pm$2.5)} ({\color{ForestGreen}+3.6}) \\
& &  & \cmark &  & 64.4 {\scriptsize($\pm$1.0)} ({\color{ForestGreen}+1.2}) & \textbf{74.5 {\scriptsize($\pm$3.6)}} ({\color{ForestGreen}+5.7}) \\
& &  &  & \cmark & 64.7 {\scriptsize($\pm$0.4)} ({\color{ForestGreen}+1.5})  & 71.1 {\scriptsize($\pm$3.1)} ({\color{ForestGreen}+2.3}) \\ \midrule
\parbox[t]{2mm}{\multirow{4}{*}{\rotatebox[origin=c]{90}{\footnotesize \specialcell{Multiple\\contexts\\}}}} & \multirow{4}{*}{\textbf{\model}} & \cmark & \cmark &  & \textbf{65.4 {\scriptsize($\pm$0.4)}} ({\color{ForestGreen}+2.2}) & 70.4 {\scriptsize($\pm$2.1)} ({\color{ForestGreen}+1.6}) \\
& & \cmark &  & \cmark  & \underline{64.9 {\scriptsize($\pm$1.6)}} ({\color{ForestGreen}+1.7}) & 73.7 {\scriptsize($\pm$3.8)} ({\color{ForestGreen}+4.9}) \\
& &  & \cmark & \cmark & 63.0 {\scriptsize($\pm$2.1)} ({\color{BrickRed}-0.2}) & \underline{74.2 {\scriptsize($\pm$2.2)}} ({\color{ForestGreen}+5.4}) \\
& & \cmark & \cmark & \cmark & 65.0 {\scriptsize($\pm$1.6)} ({\color{ForestGreen}+1.8}) & 71.5 {\scriptsize($\pm$2.8)} ({\color{ForestGreen}+2.7}) \\ \bottomrule
\end{tabular}
}
\label{tab:main_results}
\end{table*}

Snapshot Mountain Zebra comprises camera trap images taken at the Mountain Zebra National Park in South Africa as a part of the Snapshot Safari project \cite{pardo2021snapshot}.
% The Snapshot Mountain Zebra dataset, part of the Snapshot Safari project \cite{pardo2021snapshot}, consists of camera trap images captured at the Mountain Zebra National Park, South Africa.
The label space consists of 53 species, mostly annotated at the species level. Prominent animal species include Cape Mountain zebra, kudu, and springbok. The location coordinates are not available for this dataset due to privacy and security reasons. We manually split the images to have disjoint camera traps in each split due to the absence of a standard split. These datasets pose a significant challenge for species recognition due to factors like inadequate illumination, motion blur, occlusion, temporal variations, and diverse weather conditions, effectively reflecting the complexities of real-life camera trap usage.
Dataset statistics are shown in Table~\ref{tab:dataset_stat}.

\subsection{Baselines}
% {\color{blue} Add Resnet, other baselines from WILDS paper.}
We use the \model with no context that uses just the species label edges as our baseline. In addition, we compare with the following baseline algorithms for OOD generalization: Empirical Risk Minimization (ERM) \cite{DBLP:conf/icml/KohSMXZBHYPGLDS21}, which trains the model to minimize average training loss, CORAL \cite{DBLP:conf/eccv/SunS16}, a method for unsupervised domain adaptation that learns domain invariant features, Group DRO \cite{DBLP:conf/icml/HuNSS18}, an algorithm that uses distributionally robust optimization to perform well on subpopulation shifts, Fish \cite{shi2022gradient} that attempts domain adaptation using gradient matching, and ABSGD \cite{qi2020attentional}, an optimization method for addressing data imbalance.
As an alternative way of incorporating contextual information, we implement MLP-concat, a baseline which utilizes the location and temporal features at both training and inference time. It uses vanilla concatenation to fuse visual and spatiotemporal representations which are then fed into an MLP. The missing features are substituted by a mean value computed over the training dataset.
All models use a pre-trained ResNet-50 as image encoder. We evaluate the models using overall accuracy as the metric.

\subsection{Implementation Details}
% {\color{blue} Describe training strategy and hyper-parameter details}
We implement our models in PyTorch.
The hidden dimension of the multimodal KG embedding model is set to 512, with a batch size of 16.
The images are resized to $448\times 448$ before input to the image encoder.
For the location and time attributes, we use a 3-layer MLP that projects the feature input dimension to the embedding dimension and uses PReLU as the activation function.
We use Adam \cite{kingma2014adam} optimizer with a learning rate of 3e-5 and 1e-3 for the image encoder and the rest of the parameters, respectively.
% A more comprehensive description of the hyperparameter setup is given in the Appendix.
% The hidden dimension of the multimodal KG embedding model is set to 512, with a batch size of 16.
% The number of training epochs is 12 and 15 for \dataset and Snapshot Mountain Zebra, respectively.
In our experiments, the models on \dataset and Snapshot Mountain Zebra were trained for 12 and 15 epochs, respectively.
% All images are resized to $448\times 448$.
We use early stopping based on validation accuracy to prevent overfitting.
The early stopping patience parameter is set to 5 epochs.
All results are reported with averages across three random seeds.

\begin{table}[htbp]
\centering
\caption{Dataset Statistics.}
\small
\resizebox{\linewidth}{!}{%
\begin{tabular}{llcc} \toprule
\bfseries Dataset & \bfseries Split      & \bfseries \# Images & \bfseries \# Camera traps \\ \midrule
\multirow{3}{*}{iWildCam2020-WILDS} & Train      & 129,809       & 243                 \\
& Val.\ & 14,961        & 32                  \\
& Test       & 42,791        & 48                 \\ \midrule
\multirow{3}{*}{Snapshot Mountain Zebra} & Train      & 39,820       & 13                 \\
& Val.\ & 14,754        & 3                  \\
& Test       & 18,417        & 3                 \\ \bottomrule
\end{tabular}
}
\label{tab:dataset_stat}
\end{table}

% \begin{table}[htbp]
% \centering
% \small
% \caption{Dataset Statistics}
% \label{tab:dataset_stat}
% % \resizebox{\linewidth}{!}{%
% \begin{tabular}{ccccccc}
% \toprule
% \bfseries Dataset   & \multicolumn{3}{c}{\# \bfseries Images}  & \multicolumn{3}{c}{\# \bfseries No.\ camera traps} \\ \cmidrule{2-4} \cmidrule{5-7}
%           &        \bfseries Train    & \bfseries Valid   & \bfseries Test & \bfseries Train    & \bfseries Valid   & \bfseries Test                      \\ \midrule
% iWildCam2020-WILDS  & 129,809  & 14,961  & 42,791            \\
% Snapshot Mountain Zebra  & 39,820   & 14,754   & 18,417       \\ \bottomrule        
% \end{tabular}
% \end{table}

\begin{table}[htbp]
\centering
\caption{Species classification results on Snapshot Mountain Zebra dataset. We obtain the results for OOD baselines by training them on this dataset using publicly available code. We mark the improvements over \model (no-context) in {\color{ForestGreen}green}.}
\small
\resizebox{\linewidth}{!}{%
\begin{tabular}{lccl} \toprule
 \multirow{2}{*}{\bfseries Model}               &  \multicolumn{2}{c}{\bfseries Multi-modality} & \multirow{2}{*}{\bfseries Test Acc.\ (\%)}                                 \\ \cmidrule(r){2-3} 
   &  \bfseries Taxonomy & \bfseries Time                  &                      \\ \cmidrule(r){1-1} \cmidrule(r){2-3} \cmidrule(r){4-4} 
 ERM \cite{DBLP:conf/icml/KohSMXZBHYPGLDS21}               & \multicolumn{2}{c}{\multirow{4}{*}{--}}  & 96.2 {\scriptsize($\pm$0.6)}      \\
 CORAL \cite{DBLP:conf/eccv/SunS16} &  &   & 96.6 {\scriptsize($\pm$1.2)}     \\
 Group DRO \cite{DBLP:conf/icml/HuNSS18} &  &   & 93.4 {\scriptsize($\pm$2.1)} \\
 ABSGD \cite{qi2020attentional} &  &                         & 93.4 {\scriptsize($\pm$2.0)} \\
 MLP-concat &  & \cmark &  94.7 {\scriptsize($\pm$0.0)}\\ \midrule
 \model(no-context) & \multicolumn{2}{c}{--} & 92.9 {\scriptsize($\pm$2.5)} \\ \midrule
 \multirow{3}{*}{\textbf{\model}} & \cmark &  & 93.9 {\scriptsize($\pm$2.8)} ({\color{ForestGreen}+1.0}) \\
 &  & \cmark   & 95.3 {\scriptsize($\pm$3.1)} ({\color{ForestGreen}+2.4}) \\
 & \cmark & \cmark  & \textbf{96.8 {\scriptsize($\pm$0.4)}} ({\color{ForestGreen}+3.9}) \\ \bottomrule
\end{tabular}
}
\label{tab:lila_results}
\end{table}

% \newpage
\section{Results}
In this section, we attempt to answer the following questions:
\begin{enumerate}
    \item [Q1.] Does the use of contextual information contribute toward better performance? (Section~\ref{sec:main_results})
    \item [Q2.] How does \model's performance compare to the existing state-of-the-art? (Section~\ref{sec:sota_comparison})
    \item [Q3.] Does the taxonomy-aware \model model result in more semantically plausible predictions? (Section~\ref{sec:tax_case_study})
    \item [Q4.] How does \model's performance compare to baselines for  under-represented species? (Section~\ref{sec:underrep_species})
\end{enumerate}

\subsection{Performance Comparison with Addition of Multimodal Context} \label{sec:main_results}

We add taxonomy, location, and temporal context information to the base KG and observe the impact on the species classification performance. 
Table~\ref{tab:main_results} shows the results for the iWildCam2020-WILDS dataset. We make the following observations from these results:

Firstly, the addition of one or more contexts results in a performance gain over the no-context baseline in the vast majority of cases.
For instance, in the case of \model with taxonomy, we obtain a \num{3.6}\% improvement over the no-context baseline in terms of test accuracy.
Incorporating location context produces a notable \num{5.7}\% enhancement in test set accuracy, underlining the significance of auxiliary information for improved out-of-domain generalization.
We further analyze the role of location in predicting the species distribution in Section~\ref{sec:loc_time_analysis}. 
Additionally, utilizing the time attribute yields a substantial improvement over the no-context baseline, resulting in a \num{2.3}\% performance gain.

Secondly, we observe that the use of multiple contexts results in a performance boost in a majority of cases.
For instance, the addition of location and time attributes improves over the taxonomy baseline by a margin of \num{2.6}\% and \num{2.1}\% respectively in terms of the validation set accuracy.
Similarly, the taxonomy with time baseline obtains an improvement of \num{1.3}\% and \num{2.6}\% over the taxonomy and time baselines, respectively in terms of test accuracy.

Table~\ref{tab:lila_results} shows the results for the Snapshot Mountain Zebra dataset. Incorporating taxonomy and time contexts results in a performance boost of \num{1}\% and \num{2.4}\% respectively, over the no-context baseline. Furthermore, their combined use yields a noteworthy 3.9\% gain in test accuracy.

\subsection{Comparison with OOD Generalization Approaches} \label{sec:sota_comparison}
We compare the performance of the \model with methods specifically designed for out-of-domain generalization.
% shows the comparison of \model with these baseline approaches.
Notably, our best-performing model, which uses location as context, achieves state-of-the-art performance in terms of OOD test accuracy, outperforming the existing SOTA model (CORAL) by \num{1.2}\% on the iWildCam2020-WILDS dataset. Likewise, \model with taxonomy and time contexts outperforms existing approaches on the Snapshot Mountain Zebra dataset.
This demonstrates the effectiveness of leveraging diverse multimodal contexts for achieving more robust OOD generalization, even in the absence of sophisticated objectives aimed at improving domain generalization, \eg, CORAL, Group DRO, ABSGD, and Fish.
The MLP-concat baseline overfits the training camera trap locations on the iWildCam2020-WILDS dataset, resulting in suboptimal performance. \model consistently outperforms the MLP-concat baseline by a significant margin across both datasets.

\begin{table}[htbp]
\centering
\caption{Performance comparison among different KGE models (\dataset). DistMult outperforms ConvE in a majority of cases.}
\small
\begin{tabular}{lcc}
\toprule
\bfseries Model Configuration & \bfseries KGE & \bfseries Val. Acc. \\
\midrule
\multirow{2}{*}{No-context} & DistMult & \textbf{63.2 (0.4)} \\
 & ConvE    & 62.2 (0.4) \\
\midrule
\multirow{2}{*}{Taxonomy Only} & DistMult & 62.8 (2.2) \\
 & ConvE    & \textbf{64.7 (0.6)} \\
\midrule
\multirow{2}{*}{Location Only} & DistMult & \textbf{64.4 (1.0)} \\
 & ConvE    & 63.8 (1.8) \\
\midrule
\multirow{2}{*}{Time Only} & DistMult & \textbf{64.7 (0.4)} \\
 & ConvE    & 59.6 (2.8) \\
\midrule
\multirow{2}{*}{Taxonomy, Location, Time} & DistMult & \textbf{65.0 (1.6)} \\
 & ConvE    & 55.8 (3.5) \\
\bottomrule
\end{tabular}
\label{tab:distmult_conve_tbl}
\end{table}

\begin{figure*}[tbp]
    \centering
    \includegraphics[width=0.8\linewidth]{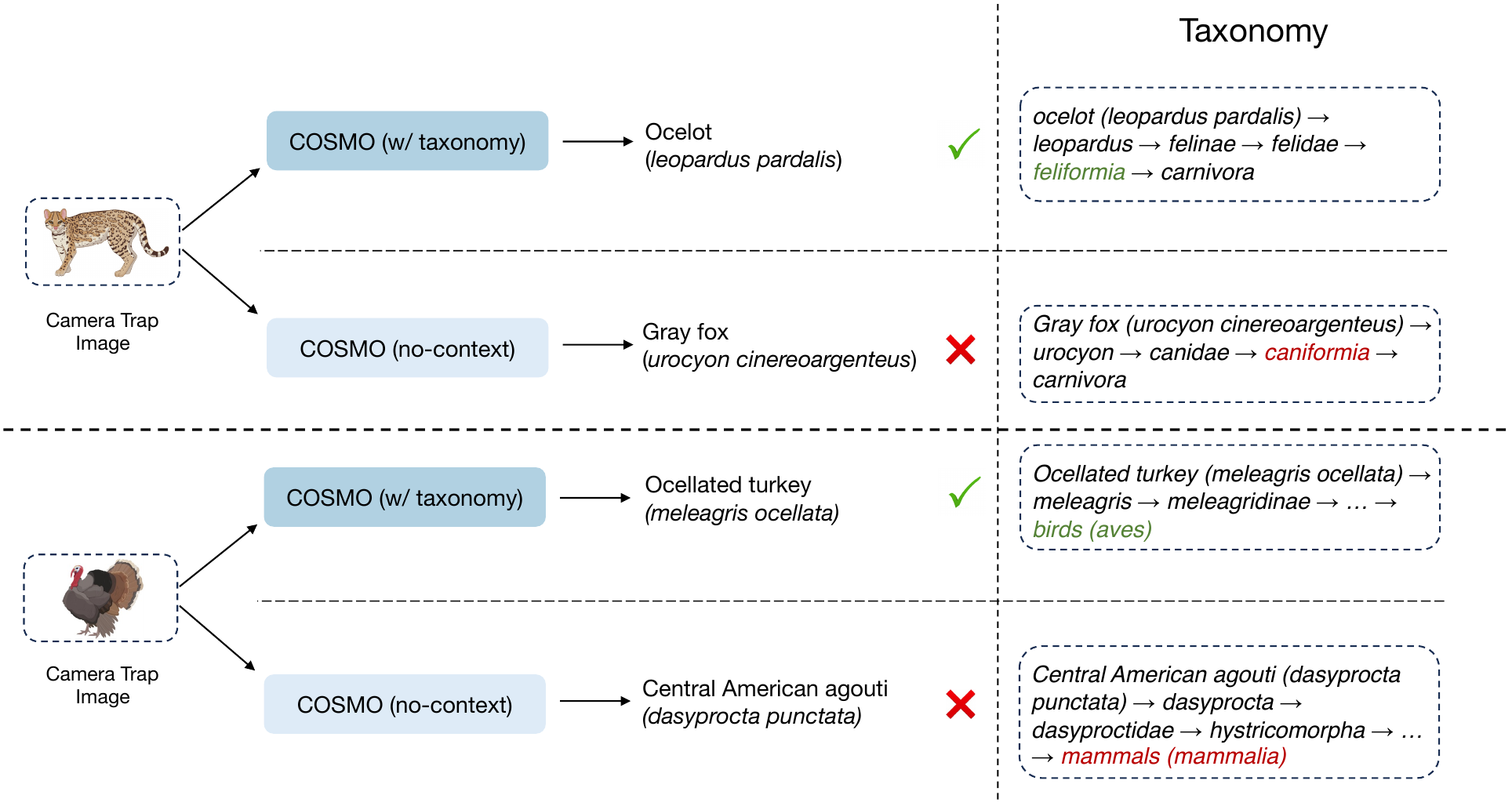}
    \caption{Comparison of \model model with and without taxonomy edges (\dataset validation set). The use of taxonomy information helps the model to avoid semantically implausible predictions.}
    \label{fig:tax}
\end{figure*}
\begin{table}[htbp]
\centering
\caption{Quantitative evaluation of \model errors with and without taxonomy using a hierarchical distance metric (\dataset). The taxonomy-aware model achieves better performance in terms of Avg.\ LCA height.}
\begin{tabular}{lc} \toprule
\bfseries Model                               & \bfseries Avg. LCA height \\ \midrule
\model (no-context)            & 6.72            \\
\textbf{\model (w/ taxonomy)} & \textbf{6.39}           \\ \bottomrule
\end{tabular}
\label{tab:taxonomy_lca}
\end{table}

\subsection{Compairson with Alternative KGE Backbones}
In our preliminary experiments, we explored the use of ConvE \cite{dettmers2018convolutional}, a strong neural network baseline, as an alternative to DistMult for the KGE backbone model for the iWildCam2020-WILDS dataset (Table~\ref{tab:distmult_conve_tbl}). Sun et al. \cite{sun-etal-2020-evaluation} show that ConvE outperforms more recent neural network KGE models when evaluated properly.
% We use ConvE \cite{dettmers2018convolutional}, a strong neural network baseline as the KGE model 
 We observe that DistMult outperforms ConvE in a majority of cases, particularly when all incorporating all context types simultaneously. 
% Therefore, we use DistMult as the backbone KGE model in our experiments.
Furthermore, DistMult offers the advantage of being more computationally efficient than neural network based KG embedding approaches.

\subsection{Fine-grained Analyses} \label{sec:fine_analysis_tax}

\subsubsection{Error Analysis for the Taxonomy-aware Model} \label{sec:tax_case_study}
To analyze the predictions of our model with and without taxonomy information, we employ a metric that takes into account the hierarchical structure of the labels.
Conventional measures like top-1 accuracy treat all errors equally, disregarding the semantic relationships among labels.
Hence, we use the Least Common Ancestor (LCA) \cite{bertinetto2020making} for the misclassified examples as the metric for this analysis (Table~\ref{tab:taxonomy_lca}). 
A lower LCA value indicates that the errors made by the taxonomy-aware model are more semantically related to the true label compared to the baseline.

We compare the predictions of \model which uses taxonomy to the no-context baseline (Figure~\ref{fig:tax}). 
Notably, the inclusion of taxonomy information assists the model in avoiding implausible predictions.
For instance, consider the case of the animal ocelot (\textit{Leopardus pardalis}), which belongs to the cat family (\textit{feliformia}). 
The use of taxonomy information prevents the misprediction of this animal as a gray fox, which belongs to the dog family (\textit{caniformia}). 
Similarly, in the second example, the baseline model incorrectly predicts the given image as Central American agouti, a mammal, instead of ocellated turkey, a bird.

% \begin{figure}[!ht]
% \centering
% \begin{subfigure}[b]{0.45\textwidth}
%     \centering
%     \includegraphics[width=0.8\linewidth]{figures/loc_corr_analysis.png}
%     \caption[Caption without FN]{Each color square shows the distance between the corresponding validation cluster centroid on x-axis and the training cluster centroid on y-axis. The correlation peaks along the diagonal\footnotemark.}
%     \label{fig:loc_corr_analysis}
% \end{subfigure}
% % \hfill
% \begin{subfigure}[b]{0.45\textwidth}
%     \centering
%     \includegraphics[width=0.8\linewidth]{figures/time_corr_analysis.png}
%     \caption{Each color square shows the distance between the corresponding training hour slot on x-axis and validation hour slot on y-axis. The correlation peaks for day-day and night-night hour slots.}
%     \label{fig:time_corr_analysis}
% \end{subfigure}
% \caption{Correlation analysis for location and time attributes. Best viewed in color.}
% \end{figure}

\begin{figure}[!ht]
\centering
\begin{subfigure}[t]{0.45\columnwidth}
    \centering
    \includegraphics[width=\linewidth]{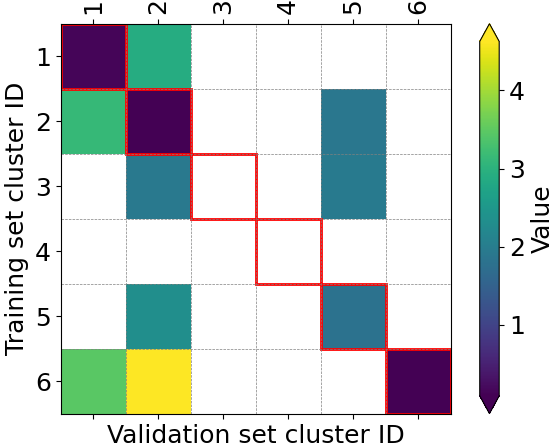}
    \caption[Caption without FN]{Each color square shows the distance between the corresponding validation cluster centroid on x-axis and the training cluster centroid on y-axis. The correlation peaks along the diagonal (highlighted in red)\footnotemark.}
    \label{fig:loc_corr_analysis}
\end{subfigure}\hfill
\begin{subfigure}[t]{0.45\columnwidth}
    \centering
    \includegraphics[width=\linewidth]{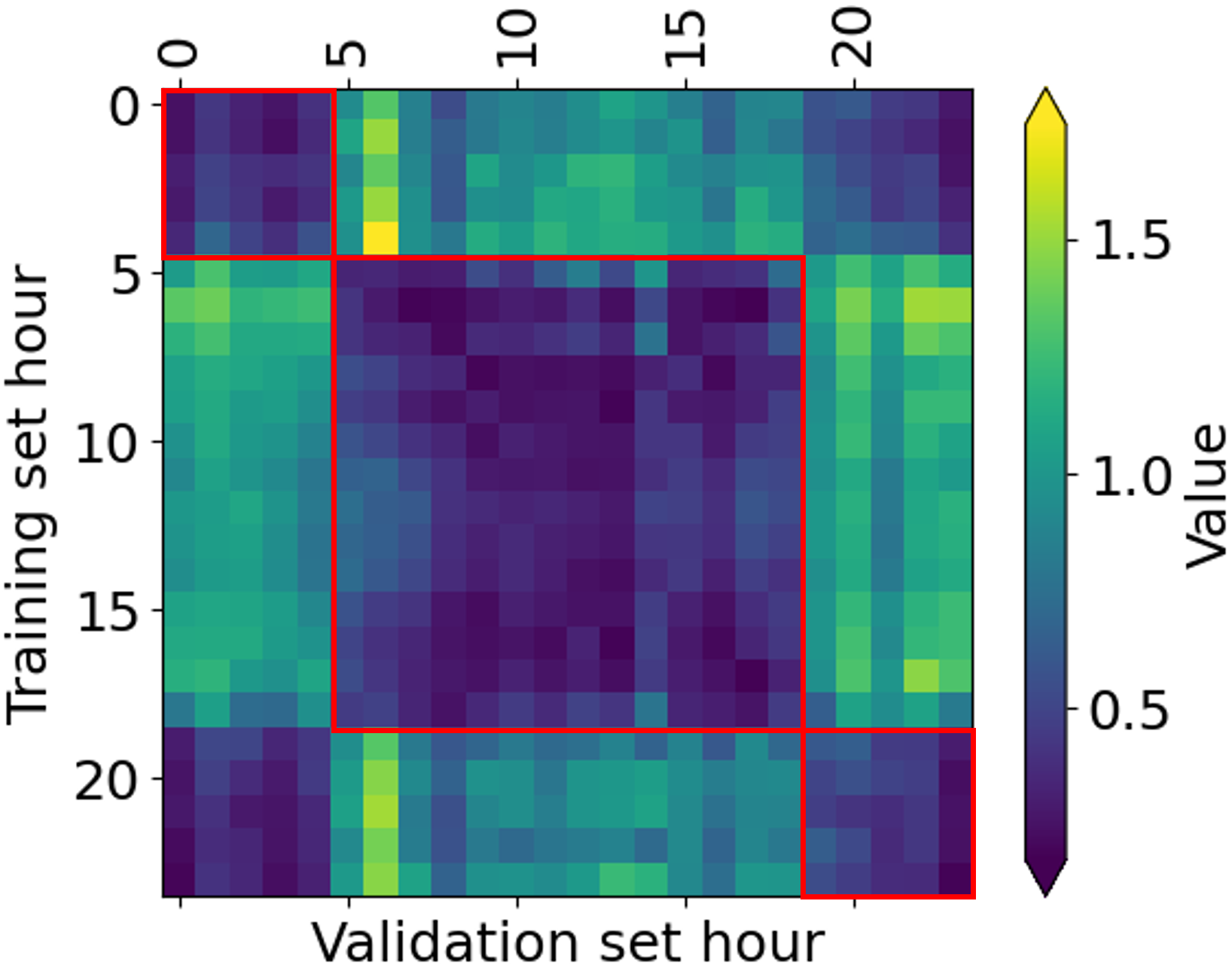}
    \caption{Each color square shows the distance between the corresponding training hour slot on x-axis and validation hour slot on y-axis. The correlation peaks for day-day and night-night hour slots (highlighted in red).}
    \label{fig:time_corr_analysis}
\end{subfigure}
\caption{Correlation analysis for location and time attributes. Best viewed in color.}
\end{figure}
\begin{figure}[htbp]
    \centering
    \includegraphics[width=0.9\linewidth]{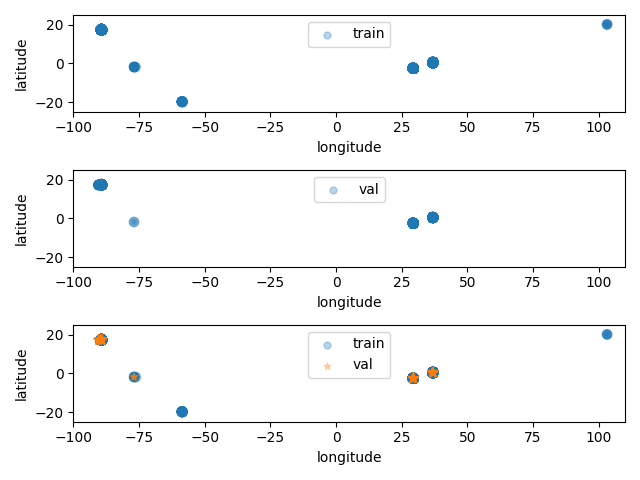}
    \caption{Plot of location GPS coordinates for training and validation splits (\dataset). The coordinates can be grouped into six clusters. Most coordinates exhibit an overlap with their respective cluster centroids at this visualization scale. Best viewed in color.}
    \label{fig:loc_split}
\end{figure}

\subsubsection{Correlation Analysis for Location and Time Attributes} \label{sec:loc_time_analysis}
We examined the relationship between species distribution and numerical attributes, such as location and time, to gain insights into how these contexts contribute to the task. The location coordinates can be grouped into six clusters. A visualization of the location clusters is shown in Figure~\ref{fig:loc_split}. For each pair of cluster centroids, we compute the Bhattacharyya distance \cite{bhattacharyya1946measure}, a measure of similarity between probability distributions, between the training and validation set species distributions (Figure~\ref{fig:loc_corr_analysis}).
% The visualization is shown in Figure~\ref{fig:loc_corr_analysis}.
Similarly, we plot the distance between species distributions corresponding to each hour of the day (Figure~\ref{fig:time_corr_analysis}).
We observe that the similarity (corresponds to lower distance) peaks along the diagonal for the location attribute, as well as for the day/night categorization of the time attribute.
This suggests these metadata give a prior for species class distribution.

\footnotetext{The null value in row 4 is due to the absence of species overlap with respective validation clusters. The null value in columns 3 and 4 indicates the absence of these clusters in the validation set.}

\begin{table}[htbp]
\centering
\caption{Performance comparison on under-represented species classification (OOD test set). The best performing \model model improves over the ERM baseline by a significant margin.}
\begin{tabular}{ll} \toprule
\bfseries Model                          & \bfseries Accuracy \\ \midrule
ERM                & 16.3    \\
\textbf{\model} & \textbf{19.0} ({\color{ForestGreen}+2.7})   \\ \bottomrule
\end{tabular}
\label{tab:underrep_species}
\end{table}

\subsubsection{Performance Comparison for Under-represented Species} \label{sec:underrep_species}

The iWildCam2020-WILDS dataset exhibits a long-tail species distribution \cite{DBLP:conf/icml/KohSMXZBHYPGLDS21}, posing challenges for accurately recognizing species that are under-represented in the training set. 
We compare the performance of our best-performing model (\model with location context) to the baseline Empirical Risk Minimization (ResNet-50) model (Table~\ref{tab:underrep_species}). We focus on examples whose labels have a maximum of 100 instances in the training set and report the overall test set accuracy for this subset of species. This selection includes species like banded palm civet, Brazilian cottontail, and leopard, all classified as vulnerable in IUCN's list of threatened species \cite{vie2009iucn}. We observe that \model outperforms the ERM baseline by \num{2.7}\%, which is a \num{16.6}\% relative improvement. These findings illustrate the potential of our model to mitigate sample inefficiency in existing approaches for under-represented species by utilizing multimodal context information.

% \FloatBarrier
\section{Discussion and Conclusion}
% In this work, we presented a novel framework in which the species classification task is reformulated as link prediction in a multimodal KG of species images and their associated heterogeneous contexts. This enables a unified way to leverage various forms of multimodal context, \eg, numerical, categorical, and taxonomy information associated with images for species identification in camera traps. Through our experiments, we demonstrate that our framework achieves superior out-of-distribution (OOD) generalization and competitive performance with state-of-the-art for species classification on the iWildCam2020-WILDS and Snapshot Mountain Zebra datasets. Additionally, our framework demonstrates improved sample efficiency in recognizing under-represented and vulnerable wildlife species. Future work will focus on integrating a more comprehensive spectrum of diverse contexts for use with real-world camera trap deployments.

In this work, we presented a novel framework in which the species classification task is reformulated as link prediction in a multimodal KG of species images and their diverse contextual information. This enables a unified way to leverage various forms of multimodal context, \eg, numerical, categorical, and taxonomy information associated with images for species classification in camera traps. Through our experiments, we demonstrate that our framework achieves superior out-of-distribution generalization and competitive performance with state-of-the-art for species classification on the iWildCam2020-WILDS and Snapshot Mountain Zebra datasets. Additionally, our framework exhibits improved sample efficiency in recognizing under-represented and vulnerable wildlife species.

% limitations
We assume that there is a perfect linkage between these contexts and the corresponding images in the training set.
However, in scenarios where such linkage is unavailable, the training procedure may introduce noise, which could lead to inferior generalization capabilities in the model.
Additionally, it is important to note that the effectiveness of diverse contexts varies based on their informativeness for the given task. 
Interestingly, combining two or more contexts could degrade performance compared to using a single context type in some cases (Table~\ref{tab:main_results}).
We posit that specific metadata, like location, might have a stronger regularization effect on improving generalization in species recognition tasks than other metadata.
To address this, future work will involve enabling the model to assign greater importance to more informative metadata.

% future directions
Furthermore, we are interested in training a foundation model for camera trap species classification across a wider spectrum of species.
% This model is expected to generalize more effectively to novel camera trap deployments across the world.
This model should demonstrate enhanced generalization capabilities for new camera trap setups worldwide.
Additionally, we aim to integrate a broader spectrum of diverse contexts such as temperature, weather conditions, habitat, and sequence information for use with real-world camera trap deployments.

\section*{Acknowledgements}
The authors would like to thank colleagues from the OSU NLP group for their constructive feedback. This research was sponsored in part by NSF OAC 2112606, NSF OAC 2118240, Cisco, and the Ohio Supercomputer Center \cite{OhioSupercomputerCenter1987}.

\newpage

\bibliographystyle{ACM-Reference-Format}
\balance
\bibliography{ref}

%%%%%%%%%%%%%%%%%%%%%%%%%%%%%%%%%%%%%%%%%%%%%%%%%%%%%%%%%%%%
\newpage
\appendix

\appendix
% \newpage
\clearpage

\section*{Supplementary Material}

\setcounter{table}{0}
\renewcommand\thetable{\Alph{section}.\arabic{table}}
\setcounter{figure}{0}
\renewcommand\thefigure{\Alph{section}.\arabic{figure}}

\section{Datasets}
% {\color{blue}TODO: include license information for iwildcam and OTT.}\\
The iWildCam2020-WILDS dataset \cite{DBLP:conf/icml/KohSMXZBHYPGLDS21} and Snapshot Mountain Zebra dataset \cite{pardo2021snapshot} are released under the Community Data License Agreement (CDLA). The Open Tree Taxonomy \cite{OTT} is licensed under  Creative Commons Attribution 1.0 Generic license. Both licenses permit their use for academic research in their original form. The \dataset label space consists of 182 species. Out of these 182 species, 155 species have support in the OTT taxonomy.

\section{Experiment Details}
Table~\ref{tab:running_time_wildcam} and Table~\ref{tab:running_time_mz} show the training time of COSMO under different settings for \dataset and Snapshot Mountain Zebra, respectively. We performed all experiments using a single Nvidia RTX A6000 GPU.

\begin{table}[H]
\centering
\caption{Training time for different ablations of COSMO (\dataset). All experiments use a single Nvidia RTX A6000 GPU.}
\small
\begin{tabular}{cccc} \toprule
\multicolumn{3}{c}{\bfseries Multi-modality}  & \multirow{2}{*}{\bfseries Running time (hrs.)}                   \\ \cmidrule(r){1-3}
\bfseries Taxonomy & \bfseries Location & \bfseries Time                  &  \\ \cmidrule(r){1-3} \cmidrule{4-4}
&  -- &           & 3.9 \\
\cmark &  &  & 3.9 \\
 & \cmark &  &  6.3\\
 &  & \cmark &  7.5\\
\cmark & \cmark &  & 6.3\\
\cmark &  & \cmark & 7.5\\
& \cmark &  \cmark & 9.9\\
\cmark & \cmark & \cmark & 9.9\\
\bottomrule
\end{tabular}
\label{tab:running_time_wildcam}
\end{table}

% \FloatBarrier

\begin{table}[H]
\centering
\caption{Training time for different ablations of COSMO (Snapshot Mountain Zebra). All experiments use a single Nvidia RTX A6000 GPU.}
\small
\begin{tabular}{ccc} \toprule
\multicolumn{2}{c}{\bfseries Multi-modality}  &  \multirow{2}{*}{\bfseries Running time (hrs.)}                  \\ \cmidrule(r){1-2}
\bfseries Taxonomy & \bfseries Time                &  \\ \cmidrule(r){1-2} \cmidrule{3-3}
\multicolumn{2}{c}{--}            &  4.2\\
\cmark  &  &  4.2\\
 & \cmark  &  7.0\\
\cmark & \cmark  & 7.0\\
\bottomrule
\end{tabular}
\label{tab:running_time_mz}
\end{table}

\section{Time Attribute Analysis}
\begin{figure}[htbp]
    \centering
    \includegraphics[width=0.9\linewidth]{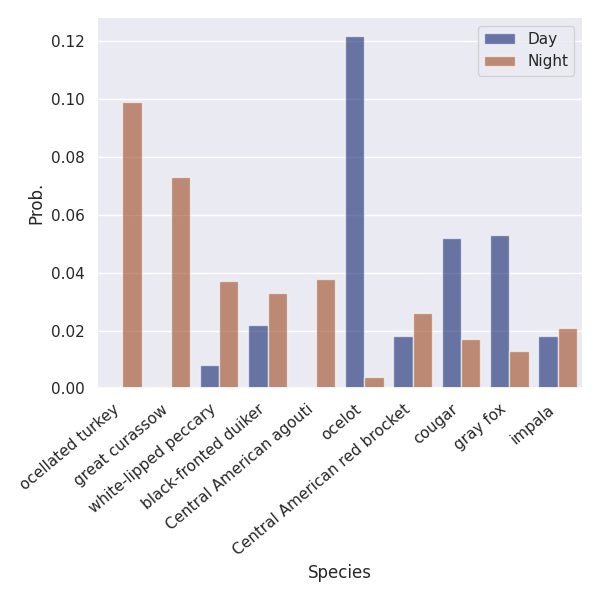}
    \caption{Species probabilities conditioned on day/night for the 10 most frequent species in the training set (\dataset). Animal species demonstrate distinct temporal preferences for their daily activities, as evidenced by the contrasting probabilities observed during day and night.}
    \label{fig:time_bar_plot}
\end{figure}

We analyzed the time attributes associated with the images of the 10 most frequent species in the training set of \dataset dataset (Figure~\ref{fig:time_bar_plot}). We observed a significant difference in the species distributions between day and night\footnote{In this analysis, we define the time duration between 5 A.M.\@ and 7 P.M.\@ local time as daytime.}. This indicates that temporal information plays a crucial role in recognizing these species.

\end{document}